%% file: main.tex
\documentclass{article} 
\usepackage{iclr2024_conference,times}

\input{math_commands.tex}

\definecolor{citecolor}{HTML}{0071BC}
\definecolor{linkcolor}{HTML}{ED1C24}
\usepackage[pagebackref=true,breaklinks=true,colorlinks,citecolor=citecolor,linkcolor=linkcolor,bookmarks=false]{hyperref}

\usepackage{url}
\usepackage{graphicx}
\usepackage{amsmath}
\usepackage{amssymb}
\usepackage{algorithm}
\usepackage{algpseudocode}
\usepackage{wrapfig}
\usepackage{booktabs}
\usepackage{enumitem}

\title{Adaptive Sharpness-Aware Pruning for \\Robust Sparse Networks}

\author{Anna Bair \thanks{Work done during an internship at NVIDIA.} \\
Carnegie Mellon University \\
\texttt{abair@cmu.edu}
\And
Hongxu Yin, Maying Shen, Pavlo Molchanov, Jose Alvarez \\
NVIDIA \\
\texttt{\{dannyy, mshen, pmolchanov, josea\}@nvidia.com}
}

\iclrfinalcopy 
\begin{document}

\maketitle

\begin{abstract}

Robustness and compactness are two essential attributes of deep learning models that are deployed in the real world. 
The goals of robustness and compactness may seem to be at odds, since robustness requires generalization across domains, while the process of compression exploits specificity in one domain. 
We introduce \textit{Adaptive Sharpness-Aware Pruning (AdaSAP)}, which unifies these goals through the lens of network sharpness. 
The AdaSAP method produces sparse networks that are robust to input variations which are \textit{unseen at training time}. 
We achieve this by strategically incorporating weight perturbations in order to optimize the loss landscape. This allows the model to be both primed for pruning and regularized for improved robustness. 
AdaSAP improves the robust accuracy of pruned models on image classification by up to +6\% on ImageNet C and +4\% on ImageNet V2, and on object detection by +4\% on a corrupted Pascal VOC dataset, over a wide range of compression ratios, pruning criteria, and network architectures, outperforming recent pruning art by large margins.

\end{abstract}

\section{Introduction}
\input{iclr_2024/intro}

\section{Related Works}
\input{iclr_2024/related_works}

\section{The AdaSAP Method}
\input{iclr_2024/method}

\section{Experiments \& Results}
\label{sec:results}
\input{iclr_2024/results}

\section{Conclusion}
\input{iclr_2024/conclusion}

\newpage
\bibliography{iclr2024_conference}
\bibliographystyle{iclr2024_conference}

\newpage
\appendix
\input{iclr_2024/appendix}

\end{document}

%% file: math_commands.tex
\usepackage{amsmath,amsfonts,bm}

\def\eqref#1{equation~\ref{#1}}

\def\1{\bm{1}}

\def\eps{{\epsilon}}

\DeclareMathAlphabet{\mathsfit}{\encodingdefault}{\sfdefault}{m}{sl}
\SetMathAlphabet{\mathsfit}{bold}{\encodingdefault}{\sfdefault}{bx}{n}

\DeclareMathOperator*{\argmax}{arg\,max}

%% file: iclr_2024/intro.tex
\begin{wrapfigure}{r}{0.53\linewidth}
    \centering
    \vspace{-1em}
    \rotatebox{270}{\includegraphics[width=5.5cm]{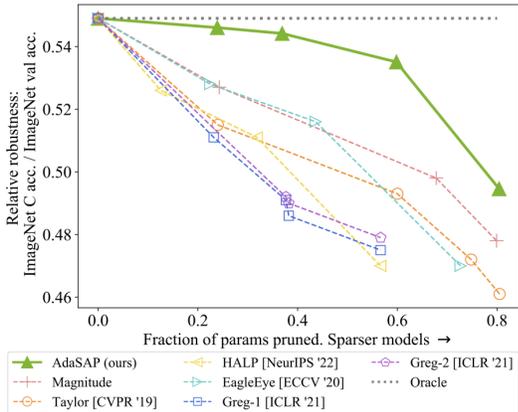}}
    \vspace{-.5em}
    \caption{Robustness of pruned models trained on ImageNet-1K drastically degrades on ImageNet-C as pruning ratio increases for many SOTA pruning methods. AdaSAP reduces the degradation in robust performance relative to standard validation performance. We approach the grey dashed line, which indicates an ideal scenario in which robust performance does not degrade at higher rates than validation performance. }
    \vspace{-1em}
    \label{fig:fig1}
\end{wrapfigure}

Deep neural networks have increasingly been used in many applications such as autonomous driving.
Unlike the controlled environments in which these models are trained, test-time inference presents new challenges including noisy real-world data and latency and memory constraints.
These challenges have led to recent efforts in network robustness and compression, but through largely separate lines of work.

Robustness to  input variation unseen during training is especially important for deployed deep learning models in safety-critical applications such as autonomous driving, where, for instance, artifacts such as dirt or snow may obscure the camera image.
More generally, in computer vision applications, this input variation falls into categories including distribution shifts, image corruptions, adversarial attacks, and label noise \citep{hendrycks2019benchmarking, azulay2018deep, carlini2016defensive, carlini2017adversarial, hendrycks2018using, steinhardt2017certified}. In attempts to address these challenges, the vision community has collected and released datasets to assess model robustness \citep{hendrycks2019benchmarking, recht2019imagenet} and investigated and improved performance on a variety of types of robustness \citep{xie2021segformer,zhou2022understanding, szegedy2013intriguing, madry2017towards}. Despite significant progress in this area, most existing robustness work focuses on dense networks \citep{guo2023robustifying, zhou2022understanding}.

\begin{figure*}[t!]
  \centering
    \includegraphics[width=\textwidth]{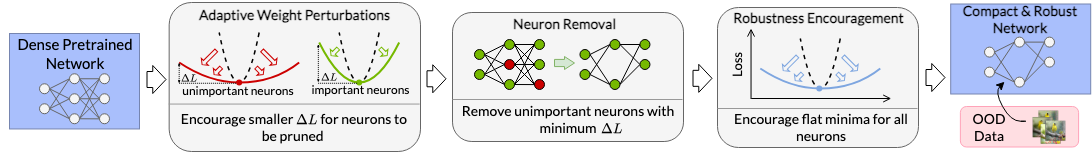}
    \vspace{-1em}
    \caption{AdaSAP is a three step process that takes as input a dense pretrained model and outputs a sparse robust model. The process can be used with any pruning method. }
    \label{fig:method_schematic}
\vspace{-0.35cm}
\end{figure*}

Running large models imposes substantial computational burdens that can be mitigated via model compression methods. 
Given the observation that neural networks often include redundant computation at inference, this line of work aims to reduce network inference costs through techniques such as 
knowledge distillation ~\citep{hinton2015distilling,molchanov2022lana,yin2020dreaming}, quantization~\citep{gholami2021survey, banner2018scalable,idelbayev2021optimal,cai2020zeroq}, pruning~\citep{han2015learning,yang2021nvit,yu2018slimmable,molchanov2016pruning, molchanov2019importance,shen2022prune}, and network adaptation~\citep{molchanov2022lana,dai2019chamnet,yang2018netadapt}.
Here we focus on pruning due to its simple procedure for network speedup.
The pruning process retains neurons deemed important over the training distribution, so it is unsurprising that pruned networks can have reduced out-of-distribution (OOD) generalization, as observed in \citet{liebenwein2021lost, hooker2019compressed}.
The effort to jointly address robustness and compression remains an open problem.

The goal of our work is to produce compact and robust neural networks. In particular, we emphasize robustness to input variations that are \textit{unseen at training time}, in contrast with methods such as robust pruning \citep{sehwag2020hydra, vemparala2021adversarial, zhao2022holistic} which assume access to the input variations at training time. It might seem as if the goals of sparsity and robustness are at odds since one aims to exploit the current dataset and task for extreme compactness whereas the other strives for maximal generalization. 
However, our method leverages a new flatness-based optimization procedure that both primes the network for pruning and improves network robustness. 
Specifically, flatness can mitigate the loss in performance during the pruning procedure and regularize the network towards improved robustness.

We introduce \textit{Adaptive Sharpness-Aware Pruning (AdaSAP)}, a three-step algorithm that prunes a network so that it is robust to OOD inputs. (1) We introduce the novel \textit{adaptive weight perturbations} which are incorporated into the optimization procedure to adaptively penalize sharpness in order to prepare the network for pruning,
(2) apply any pruning method to the model, and 
(3) continue training the model while uniformly penalizing sharpness across the network to encourage robustness. 

AdaSAP significantly improves the relative robustness over prior pruning art, seen in Figure~\ref{fig:fig1}. 
As models are pruned (with SOTA methods) more, their performance on corrupted images suffers disproportionately as compared to their validation performance. This highlights the lack of robustness preservation in recent SOTA pruning methods and AdaSAP's success at reducing this degradation\footnote{Relative robustness in Fig.~\ref{fig:fig1} refers to the ratio between the accuracy on corrupted images and the standard validation accuracy at a given prune ratio. A dense model in this setup retains $55\%$ of its validation performance when given corrupted images (location of grey dashed line). }. 

Our contributions can be summarized as follows:

\setlist[itemize]{itemsep=0pt}
\begin{itemize}[leftmargin=*]
\item We introduce \textit{AdaSAP}, a sharpness-aware pruning and fine-tuning process to jointly optimize for sparsity and robustness in deep neural networks. 
\item We propose novel \textit{adaptive weight perturbations} that prepare a network for both pruning and robustness to OOD inputs. 
This strategy actively manipulates the flatness of the loss surface so as to minimize the impact of removing neurons with lower importance scores. 
\item We demonstrate state-of-the-art performance 
compared to prior art by noticeable margins across a wide range of setups, covering (i) two tasks (classification and detection), (ii) two OOD types (image corruption and distribution shift), (iii) four networks (ResNet50 and MobileNet V1/V2 for classification, SSD512 for detection), (iv) two pruning paradigms (parameter-based and latency-based), and (v)  sparsity ratios spanning from around 20\% to 80\%. 
\item We present a detailed evaluation of robustness and compactness and show, as a first encouraging attempt, that both goals can be unified through the lens of sharpness, with analysis to encourage the community to continue pursuing this direction. 
\end{itemize}

%% file: iclr_2024/related_works.tex
\begin{figure}[t]
\begin{minipage}{0.48\textwidth}
    \centering
    \includegraphics[width=\linewidth]{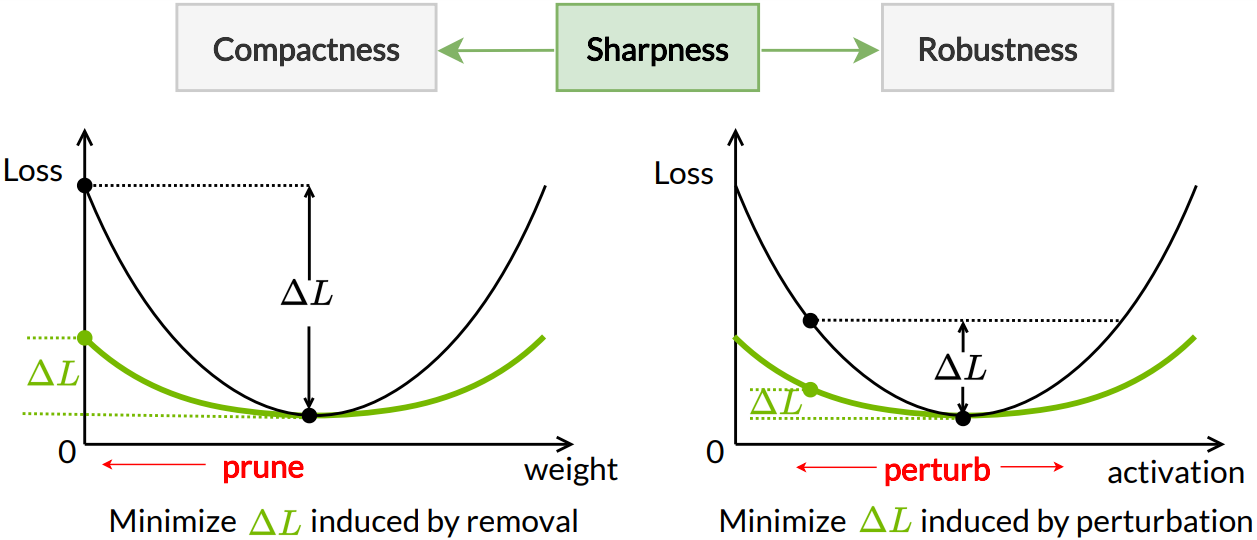}
    \vspace{-0.5em}
    \caption{\textbf{(Left)} Before pruning, encourage neurons that will be pruned to lie within a flat minimum, since their removal will affect the loss less. \textbf{(Right)} After pruning, promote robustness by encouraging flatness across the network.}
    \label{fig:flat_minima_schematic}
\end{minipage}
\hspace{0.4em}
\begin{minipage}{0.48\textwidth}
\vspace{-2em}
\begin{algorithm}[H]
\caption{AdaSAP Optimization Iteration}
\label{adasam_alg_one_epoch}
\begin{algorithmic}
\footnotesize
    \Require model weights $\mathbf{w}$ partitioned into neurons $\mathbf{w}_i$, training batch $b$, $\rho$ bounds ($\rho_{\min}$, $\rho_{\max}$), loss $L$, score function $\psi$, learning rate $\eta$
    \For{each neuron $\mathbf{w}_i$}
        \State $s_i = \psi(\mathbf{w}_i)$ \Comment{Score}
        \State Compute $\rho_i$ as in Eq. \ref{alpha_mapping}  \Comment{Determine perturbation ball size}
        \State Compute $\hat{\epsilon}_i$ as in Eq. \ref{eps_hat} \Comment{Optimal perturbation}
        \State $g_i \approx \nabla_{\mathbf{w}_i} L_{b, \mathbf{w}}(\mathbf{w}_i)|_{\mathbf{w}_i + \hat{\epsilon}_i}\ $ \Comment{Gradient approximation}
    \EndFor
    \State $\textbf{w} = \textbf{w} - \eta \textbf{g}$ \\
\Return{$\textbf{w}$}
\end{algorithmic}
\end{algorithm}
\end{minipage}
\end{figure}

\noindent
\textbf{Robustness.}
Researchers and practitioners care about many types of robustness, including adversarial robustness, robustness to distribution shift, robustness to image corruptions such as weather effects, and robustness to label noise or data poisoning \citep{szegedy2013intriguing, madry2017towards, hendrycks2019benchmarking, azulay2018deep, carlini2016defensive, carlini2017adversarial, hendrycks2018using, steinhardt2017certified}.
Several prior works examine the adversarial robustness of sparse networks \citep{sehwag2020hydra, fu2021drawing}.
Simultaneously, work such as \citet{stutz2021relating} suggests that adversarially robust networks have flatter minima.

In this work we focus on robustness to corrupted images \citep{hendrycks2019benchmarking, recht2019imagenet} due to its real-world applications such as autonomous driving. 
\citet{diffenderfer2021winning} shows that sparse subnetworks can outperform the original dense model on clean and robust accuracy. However, almost all current pruning methods still lead to worse performance on OOD data than dense networks, a result we replicate in Figure \ref{fig:fig1} \citep{liebenwein2021lost, hooker2019compressed}. 

\noindent
\textbf{Efficiency.} Efforts to improve network efficiency include pruning, distillation, quantization, and adaptation ~\citep{han2015deep, han2015learning,yu2018slimmable,molchanov2016pruning, molchanov2019importance,molchanov2022lana,shen2022prune, hinton2015distilling, gholami2021survey, banner2018scalable, dai2019chamnet,yang2018netadapt}. 
Two broad categories of pruning are structured and unstructured pruning. While unstructured pruning removes individual weights and can retain strong performance at high sparsities, structured pruning removes larger elements in the network, such as convolutional channels, and often allows for direct size or latency reduction of the pruned model~\citep{anwar2017structured, molchanov2019importance,shen2022prune}. Unstructured pruning dates back to Optimal Brain Damage, but the Lottery Ticket Hypothesis has incited recent interest in the topic and spurred developments such as SNIP and GraSP ~\citep{lee2018snip, frankle2018lottery, hoffmann2021towards, lecun1989optimal, tartaglione2022loss, wang2020picking, yu2022hessian}.
In the current work, we focus on structured channel-wise pruning.

\noindent
\textbf{Flat minima.}
The relationship between flat minima and generalization has been studied for many years \citep{hochreiter1997flat, keskar2016large, wu2017towards, guo2020expandnets, jiang2019fantastic, wu2020adversarial, bartoldson2020generalization, cha2021swad, izmailov2018averaging, kaddourflat}. 
Flat minima are often correlated with improved generalization, but the investigations into the extent of this correlation and the underlying causal relationship are still ongoing. 
Also, flatness is metric-dependent and sharp minima can generalize well \citep{dinh2017sharp}.
Sharpness-Aware Minimization (SAM) \citep{foret2020sharpness} is a way to optimize for finding flatter minima and leads to better generalization. Work which builds upon SAM improves its efficiency \citep{liu2022towards, du2022sharpness, jiang2023adaptive}, modifies the optimization for improved performance \citep{kwon2021asam}, applies it for compression in NLP applications \citep{na2022train} and as a pre-pruning optimizer \citep{zhou2023threeregime}, and modifies the objective for unstructured pruning \citep{peste2022cram}. 
Stochastic weight averaging (SWA) is another method that has been used to optimize for flatter minima \citep{izmailov2018averaging, cha2021swad, kaddourflat}.
The present work is inspired by SAM and its use in model compression \citep{foret2020sharpness, na2022train} but differs in its use of adaptive weight perturbations and its emphasis on robustness. In this work we use the formulation introduced in ASAM \citep{kwon2021asam} due to its improved performance over that of SAM.

%% file: iclr_2024/method.tex
\begin{wrapfigure}{r}{0.7\linewidth}
\begin{minipage}{\linewidth}
\vspace{-2.2em}
\begin{algorithm}[H]
\caption{AdaSAP Pruning Procedure }
\label{adasam_pruning}
\begin{algorithmic}
\footnotesize
    \Require Pretrained model weights $\mathbf{w}$ partitioned into neurons $w_i$, pruning importance criteria $\phi$, $\rho$ bounds ($\rho_{\min}$, $\rho_{\max}$)

    \State Let $\texttt{train\_iter} =$ Algorithm \ref{adasam_alg_one_epoch}
    \For{epoch \textbf{in} warmup\_epochs} \Comment{Adaptive Weight Perturbation}
        \For{each iteration}
            \State Sample training batch $b$
            \State $\mathbf{w}$ = \texttt{train\_iter} ($\mathbf{w}$, $b$, $\rho_{\min}$, $\rho_{\max}$)
        \EndFor
    \EndFor 
    
    \For{epoch \textbf{in} pruning\_epochs} \Comment{Pruning}
        \For{each iteration} 
            \State Sample training batch $b$
            \State $\mathbf{w}$ = \texttt{train\_iter} ($\mathbf{w}$, $b$, $\rho_{\min}$, $\rho_{\max}$)
            \If{iteration \% prune\_frequency = 0}
            \State $\mathbf{s} = \phi(\mathbf{w})$ \Comment{Score all neurons}
            \State $\texttt{idxs} = \texttt{rank}(s)[:\texttt{prune\_num}]$ \Comment{Neurons with lowest score}
            \State $\mathbf{w}_{\texttt{idxs}} = 0$ \Comment{Prune neurons}
            \EndIf
  
        \EndFor
    \EndFor 
    \For{epoch \textbf{in} finetune\_epochs} \Comment{Robustness encouragement}
        \For{each iteration}
            \State Sample training batch $b$
            \State $\mathbf{w}$ = \texttt{train\_iter} ($\mathbf{w}$, $b$, $\rho_{\min}=\rho_{\max}=\texttt{constant}$) \Comment{SAM}
        \EndFor
  
    \EndFor
    \\ \Return{$\mathbf{w}$}
    \Ensure Pruned and finetuned model

\end{algorithmic}
\end{algorithm}
\end{minipage}
\end{wrapfigure}

 Flat minima are minima within relatively large regions of the parameter space that have low loss.
It has previously been observed that optimizing for flatter minima improves generalization and adversarial robustness in deep networks and generalization in sparse networks~\citep{foret2020sharpness, stutz2021relating,na2022train}.
 Based on these observations, we hypothesize that optimizing for flatter minima during pruning can also enhance robustness. 
 The main objective of our design is to find flat minima in order to produce models that are simultaneously prunable and robust.
 We introduce a new method,  Adaptive Sharpness-Aware Pruning (AdaSAP). Figure \ref{fig:method_schematic} and Algorithm \ref{adasam_pruning} detail the following three step procedure:

\noindent
\textbf{(1) Adaptive Weight Perturbations}.
During warmup, we ensure that the neurons that will be pruned, are regularized to lie in flat regions so that they will not cause the model's performance to suffer too much after pruning. 

\noindent
\textbf{(2) Neuron Removal}.
We conduct structured channel pruning and remove neurons according to any pruning criteria. Here we use magnitude pruning \citep{han2015learning}. During pruning, the model will preserve much of its performance since unimportant neurons have already been situated in flatter loss regions.
    
\noindent
\textbf{(3) Robustness Encouragement}. 
While training the pruned model to convergence, we enforce flat minima, this time uniformly (\textit{i.e.}, non-adaptively) across the entire network to promote robustness. This step is compatible with any flatness-based optimizer. Here we use ASAM \citep{ kwon2021asam}.

\renewcommand{\algorithmicrequire}{\textbf{Input:}}
\renewcommand{\algorithmicensure}{\textbf{Output:}}

\subsection{Adaptive Weight Perturbation}

\textbf{Flatness-informed pruning. }Our procedure performs gradient updates which are informed by the local flatness in order to best prepare the network for pruning.
We adapt the regularization per neuron based on the importance, or likelihood of that neuron to be pruned.
Pruning can be viewed as a special case of directional robustness, in which the weight and activation of a pruned neuron are set to zero. 
When the loss landscape is flat, the pruning process will incur only a small impact on the performance of the model. 
This formulation indicates that pruning and robustness share a common characterization through the perspective of sharpness, as demonstrated in Figure \ref{fig:flat_minima_schematic}.

Taking inspiration from the uniform  $\eps$-ball perturbation based method in~\citet{foret2020sharpness} and its scale-invariant version in \citet{kwon2021asam}, among several other approaches for achieving flatness in dense models~\citep{izmailov2018averaging, santurkar2018does}, we perturb  neurons according to their importance to better prepare the network for pruning.
We follow the intuition that important weights are worth adding some sharpness to our model, while unimportant weights can have stronger regularization. This idea is supported by \citet{molchanov2019importance}, which shows that pruning neurons in flat regions hurts the loss less than pruning neurons in sharp regions.
We adapt the size of allowable perturbations based on the likelihood of each weight to be pruned: weights with low importance scores have large perturbation balls which enforce that they lie within flatter regions and vice versa.

\textbf{Setup.} Following this intuition, we derive the gradient update for training. 
Consider a network whose parameters are grouped into $I$ partitions, where each partition $i$ includes the set of weights $\mathbf{w}_i$. 
In our setting, since we consider channel-wise pruning, each channel constitutes a partition.
For simplicity, throughout this paper we refer to a channel-wise group of weights as a \textit{neuron}.
We use $\mathbf{w}_i$ to refer to a single neuron and $\textbf{w}$ to refer to the collection of all neurons $\{\mathbf{w}_i\}_{i=1}^I$.

\textbf{Perturbation regions.} Consider training loss $L_S(\mathbf{w})$ of the network over the training dataset $S$. Consider a set of weight perturbation ball radius sizes $\boldsymbol\rho$, with each $\rho_i$ corresponding to the size of allowable perturbations for neuron $\mathbf{w}_i$.
Our goal is to optimize for a network in which unimportant neurons lie in flatter minima prior to pruning via the following objective
\begin{equation}
    \min_{\mathbf{w}} \max_{\{\eps: \|T_{\mathbf{w}}^{-1}\eps_i\| \leq \mathbf{\rho}_i\}_{i=1}^I} L_S(\mathbf{w} + \mathbf{\eps}). \label{objective}
\end{equation}
where $T_{\mathbf{w}}$ and its inverse $T_{\mathbf{w}}^{-1}$  are transformations that can be applied during optimization so as to reshape the perturbation region (\textit{i.e.}, not necessarily a ball). \citet{kwon2021asam} showed that this strategy allows for greater perturbation exploration and leads to improved performance.
$\rho_i$ values are based on computing an importance score for each neuron, similar to a pruning criterion. Consider a neuron importance score function $\psi(\cdot)$, such as the $\ell_2$ norm of the neuron weight or other scores that are some function of weights, gradients, and higher order terms \citep{molchanov2019importance, lee2018snip, yu2022hessian, shen2022prune, wang2020picking}. Given each neuron score $s_i = \psi(\mathbf{w}_i)$, we can now compute the perturbation ball size $\rho_i$, which projects $s_i$ to lie within $(\rho_{min}, \rho_{max})$ 
\begin{equation}
    \rho_i = \rho_{\max} - \frac{s_i - s_{\min}}{s_{\max} - s_{\min}} (\rho_{\max} - \rho_{\min}). \label{alpha_mapping}
\end{equation}
Scores $s_{\min}$ and $s_{\max}$ are obtained empirically at each gradient step, but $\rho_{\max}$ and $\rho_{\min}$ are hyperparameters that are set ahead of time and we find them to scale across various experiment settings (further details in Section \ref{sec:results} and Appendix \ref{appendix_exp}).
Neuron importance estimates $\psi(\cdot)$ can be evaluated during training with little additional computational overhead.

\textbf{Optimal gradient update.} Our enhanced gradient update optimizes for finding a minimum that also lies in a region of low loss. 
Beginning with our objective in Eq. \ref{objective},  we perform a series of approximations in order to find the most adversarial perturbations $\boldsymbol\eps = \{\eps_i\}$ (\textit{i.e.,} highest loss near the minimum) which we can apply to the neurons $\textbf{w} = \{\mathbf{w}_i\}$ in our network in order to obtain a flatness-informed gradient update.

We use a first order Taylor expansion around each neuron $i$ with respect to $\boldsymbol\eps$ around $0$ (in line with \citet{foret2020sharpness}) to approximate the inner maximization in Eq.~\ref{objective}. This gives us $\boldsymbol\eps^*$, the optimal $\boldsymbol\eps$ we can use to inform our gradient update. Define $\tilde{\epsilon}_i = T_{\mathbf{w}}^{-1}\epsilon_i$. For simplicity we denote the conditional loss due to weight perturbations as the function $L_{S, \mathbf{w}}(\cdot)$. Then we derive our optimal update as follows for each neuron: 
\begin{align}
    \tilde{\eps}^*_i &= \argmax_{\|\tilde{\eps}_i\|_2 \leq \rho_i} L_{S, \mathbf{w}} (\mathbf{w}_i + T_{\mathbf{w}}\tilde{\eps}_i), \nonumber \\ \nonumber
    &\approx \argmax_{\|\tilde{\eps}_i\|_2 \leq \rho_i} L_{S, \mathbf{w}} (\mathbf{w}_i) + \tilde{\eps}_i^\top T_{\mathbf{w}} \nabla_{\mathbf{w}_i} L_{S, \mathbf{w}}(\mathbf{w}_i), \\
    &= \argmax_{\|\tilde{\eps}_i\|_2 \leq \rho_i} \tilde{\eps}_i^\top T_{\mathbf{w}} \nabla_{\mathbf{w}_i} L_{S, \mathbf{w}}(\mathbf{w}_i),
\end{align}
$\|\cdot\|_2$ being the $\ell_2$ norm. We can then approximate $\tilde{\boldsymbol\eps}^*$
by 
rescaling the gradient associated with each neuron so that its norm is $\rho_i$ through
\begin{equation}
    \tilde{\eps}^*_i =  \rho_i \frac{ \ \mathrm{sign} (\nabla_{w_i} L_{S, \mathbf{w}}(\mathbf{w}_i)) |T_{\mathbf{w}} \nabla_{\mathbf{w}_i} L_{S, \mathbf{w}}(\mathbf{w}_i)| }{ (\|T_{\mathbf{w}} \nabla_{\mathbf{w}_i} L_{S, \mathbf{w}}(\mathbf{w}_i)\|_2^2)^{1/2} }. 
\end{equation}
Since we were originally approximating $\tilde{\epsilon}_i = T_{\mathbf{w}}^{-1}\epsilon_i$, we recover our approximation of $\mathbf{\eps}^*$ as
\begin{equation}
    \hat{\eps}_i= \rho_i \frac{T_{\mathbf{w}}^2 \nabla_{\mathbf{w}_i} L_{S, \mathbf{w}}(\mathbf{w}_i)}{\|\nabla_{\mathbf{w}_i} L_{S, \mathbf{w}}(\mathbf{w}_i)\|_2}.
    \label{eps_hat}
\end{equation}
\textbf{Approximating the optimal gradient update.} We then find an approximation to our desired gradient. Referring to  Eq. \ref{objective}, we can find the gradient of the inner maximization in order to derive a sharpness-aware gradient update. We use the inner maximization and plug in our approximation $\hat{\eps}$ for each individual neuron during adjacent pruning steps, defined as
\begin{align}
    \nabla_{\mathbf{w}_i} \max_{\|T_{\mathbf{w}}^{-1}\eps_i\|_2 \leq \rho_i} L_{S, \mathbf{w}}(\mathbf{w}_i + \eps_i) &\approx \nabla_{\mathbf{w}_i} L_{S, \mathbf{w}}(\mathbf{w}_i + \hat{\eps}_i) \nonumber \\
    &= \frac{d(\mathbf{w}_i + \hat{\eps}_i)}{d\mathbf{w}_i} \nabla_{\mathbf{w}_i} L_{S, \mathbf{w}}(\mathbf{w}_i) |_{\mathbf{w}_i + \hat{\eps}_i} \nonumber \\
    &= \nabla_{\mathbf{w}_i} L_{S, \mathbf{w}}(\mathbf{w}_i)|_{\mathbf{w}_i + \hat{\eps}_i} \nonumber + \frac{d\hat{\eps}_i}{d\mathbf{w}_i} \nabla_{\mathbf{w}_i} L_{S, \mathbf{w}}(\mathbf{w}_i)|_{\mathbf{w}_i + \hat{\eps}_i} \nonumber \\
    &\approx \nabla_{\mathbf{w}_i} L_{S, \mathbf{w}}(\mathbf{w}_i)|_{\mathbf{w}_i + \hat{\eps}_i}, 
\end{align}
\noindent where we drop the second order terms for efficiency. 
This leads us to our final gradient update approximation between adjacent pruning steps as $\nabla_{\mathbf{w}_i} L_{S, \mathbf{w}}(\mathbf{w}_i) \approx \nabla_{\mathbf{w}_i} L_{S, \mathbf{w}}(\mathbf{w}_i)|_{\mathbf{w}_i + \hat{\eps}_i}$.

We use this gradient update in place of the standard gradient update during the warmup phase of our pruning procedure before neuron removal. This sets up our network so that when pruning begins, we have a model that is closer to a flat minima.
Optimization details are in Algorithm \ref{adasam_alg_one_epoch}.

\subsection{Neuron Removal}

We focus on structured (channel-wise) pruning since it leads to models that can take advantage of direct computational resource savings on GPUs, leading to inference speedup~\citep{molchanov2022lana,shen2022prune,yang2021nvit}.

In this stage of the procedure, we remove unimportant neurons according to any scoring function $\phi(\cdot)$ that measures neuron saliency.
This may be the same or different from the scoring function $\psi(\cdot)$ used in determining adaptive weight perturbation ball sizes during the first step.
AdaSAP works with a range of pruning methods as we show later in Section \ref{sec:results}.

\subsection{Robustness Encouragement}
In the third and final phase, we focus on optimizing our model for robust performance. 
We regularize our weights uniformly since we have completed pruning and now want to enforce robustness for the entire network.
This is illustrated in Figure \ref{fig:flat_minima_schematic} in the plot on the right: the network loss relative to a flatter neuron will stay more stable in the presence of corruptions, whereas the loss relative to a sharper neuron could fluctuate significantly. This portion of the procedure could be instantiated with any sharpness-based method that optimizes for an overall flatter minima.

\subsection{Final Comments}
We emphasize that AdaSAP \textbf{\textit{is not a pruning method}} but rather an optimization paradigm that performs robustness-aware pruning. 
Our method can be used in conjunction with any existing pruning criteria. In Section \ref{sec:results} we show that AdaSAP (in conjunction with magnitude pruning) generally outperforms existing SOTA pruning methods. As new pruning techniques arise, they could be enhanced via AdaSAP in order to obtain increased performance, particularly robust performance.

Additionally, we note that our method \textit{\textbf{differs from robust pruning methods}} such as \citet{sehwag2020hydra, vemparala2021adversarial, zhao2022holistic} since we consider a setting in which the model may be exposed to \textit{novel types of corruptions at test time}. Therefore, we do not want to include, i.e.  ImageNet C or ImageNet V2 images in our training set. We demonstrate results on these OOD datasets in an attempt to demonstrate the versatility of our method: despite only training on ImageNet data, we can still expect the models to be robust to unseen input variation.

%% file: iclr_2024/results.tex
\subsection{Experiment Details}
We describe the basics of our experimental setup here. Additional experiment details and confidence intervals can be found in Appendix \ref{appendix_exp}.

\noindent
\textbf{Datasets.} For image classification, we train on ImageNet-1K ~\citep{deng2009imagenet} 
 and additionally evaluate on ImageNet-C~\citep{hendrycks2019benchmarking} and ImageNet-V2~\citep{recht2019imagenet}.
For object detection, we use the Pascal VOC dataset \citep{everingham2009pascal}. 
To assess robustness, we create a Pascal VOC-C dataset by applying ImageNet-C-style corruptions to the test set.

\noindent
\textbf{Pruning criteria.} Our proposal is agnostic to the particular importance criteria $\phi$ used to prune the network.
We evaluate our method on two types of pruning: \textit{parameter-based} and \textit{latency-based}. 
We choose to group our methods into these two categories due to the different goals of pruning methods.
Some methods, such as magnitude pruning, seek to reduce the number of parameters to produce a small model.
Other methods, such as HALP \citep{shen2022structural}, seek to produce a model with the fastest latency speedup, regardless of the number of parameters.

In both cases, we perform structured pruning, that is, pruning channels rather than individual parameters. 
We refer to our two methods as $\text{AdaSAP}_P$ and $\text{AdaSAP}_L$, to denote the parameter-specific method and the latency-specific method, respectively.
We use $\ell_2$ norm magnitude pruning for $\text{AdaSAP}_P$ and HALP for $\text{AdaSAP}_L$. Details of the pruning schedule are in Appendix \ref{appendix_exp}.

\noindent
\textbf{Metrics.} For image classification, we report the Top1 accuracy on each dataset and two \textit{robustness ratios}, defined as the ratio in robust accuracy to validation accuracy: $R_\text{C} = \text{acc}_{\text{C}} / \text{acc}_{val}$ and $R_{\text{V2}} = \text{acc}_{\text{V2}} / \text{acc}_{val}$.
Improvement in these ratios indicates that we are closing the gap between robust and validation performance.
For object detection, we evaluate the model on Pascal VOC and VOC-C. We report mean average precision (mAP) and a robustness ratio $R_\text{C} = \text{mAP}_{\text{C}} / \text{mAP}_{val}$.

\noindent
\textbf{Baselines.}
We primarily focus on comparing to two SOTA baselines which we re-run in our experiments, Taylor pruning for parameter-based pruning~\citep{molchanov2019importance}, and HALP for latency-based pruning~\citep{shen2022prune}.
Taylor pruning assigns a score to each neuron 
based on a Taylor series approximation of how much the loss would change if the neuron were removed.
HALP is a pruning method that optimizes for improving latency.
We additionally run results on magnitude pruning \citep{han2015learning}, to demonstrate the improvements of AdaSAP$_P$ over its closest baseline.
We also compare to the results cited in several other pruning methods such as GR-eg~\citep{wang2020neural}, EagleEye \citep{li2020eagleeye}, ABCPruner~\citep{lin2020channel},  and SMCP \citep{humble2022soft}.

\subsection{ImageNet Classification}
\textbf{Parameter reduction.}
Table \ref{tab:resnet-result} shows how AdaSAP compares favorably to other pruning methods at a variety of pruning ratios for ResNet50.
MobileNet V1 and V2 results show a similar trend and are included in Appendix \ref{appendix_results}.
First, AdaSAP has the best overall $R_\text{C}$ and $R_{\text{V2}}$ ratios among all pruning methods.
This means that our method helps to close the gap between robust performance and standard validation performance. Additionally, our method consistently outperforms comparisons on both  ImageNet validation performance and on robust performance (via ImageNet C and ImageNet V2)  while at the same or even smaller compressed model size. 

\begin{table}[t]
    \centering
    \footnotesize
    \caption{\textbf{ResNet50 - Parameter Reduction.}}
        \begin{tabular}{lcccccc}
        \toprule
        Method & Size $\downarrow$ & Val & $R_\text{V2}$ & $R_\text{C}$  & IN-V2  & IN-C \\
        \midrule
        \midrule
        \multicolumn{7}{c}{\textbf{ResNet50}} \\
        Dense  & $1$ &  $77.32$  & $0.83$ & $0.55$   & $64.79$ & $42.46$\\
        \midrule
        Magnitude  & $0.20$ & $73.80$  & $0.83$ & $0.48$  & $61.40$ & $35.27$ \\
        Taylor & $0.20$ & $73.56$   & $0.82$  & $0.46$  & $60.56$ & $33.93$\\
        EagleEye 1G  & $0.27$& $74.13$  & $0.83$  & $0.47$ & $61.30$ & $34.84$ \\
        \textbf{AdaSAP$_P$}  & $\mathbf{0.20}$ & $\mathbf{74.63}$ & $\mathbf{0.83}$ & $\mathbf{0.50}$  & $\mathbf{62.08}$  & $\mathbf{37.30}$\\
        \midrule
        Magnitude & $0.44$ & $76.83$ & $\mathbf{0.85}$ & $0.51$   &$\mathbf{64.92}$ & $39.27$ \\
        Taylor & $0.42$ & $75.85$  & $0.84$ & $0.50$  & $63.51$ & $37.84$  \\
        Greg-1  & $0.43$ & $73.72$ & $0.82$ & $0.48$ & $60.47$  & $35.04$ \\
        Greg-2  & $0.43$ & $73.84$  & $0.83$ & $0.48$  & $61.05$ & $35.39$ \\
        ABCPruner  & $0.44$  & $73.52$  & $0.82$ & $0.50$  & $60.46$ & $36.64$ \\
        \textbf{AdaSAP$_P$}  & $\mathbf{0.40}$ & $\mathbf{77.27}$ & $0.83$ & $\mathbf{0.53}$  & $64.51$ & $\mathbf{41.23}$  \\
        \midrule
        Greg-1  & $0.62$ & $75.16$ & $0.82$ & $0.49$  & $61.82$ & $36.88$ \\
        Greg-2 & $0.62$ & $75.36$  & $0.82$  & $0.49$ & $62.06$ & $37.09$\\
        ABCPruner & $0.71$ & $74.84$  & $0.83$ & $0.51$ & $61.73$ & $38.35$ \\
        \textbf{AdaSAP$_P$ }   & $\mathbf{0.62}$ & $\mathbf{77.99}$ & $\mathbf{0.84}$ & $\mathbf{0.52}$ & $\mathbf{65.49}$ & $\mathbf{42.68}$ \\
        \midrule
        Magnitude  & $0.76$ & $77.32$ & $0.84$ & $0.53$  & $65.20$ & $40.73$ \\
        Taylor & $0.76$  & $77.05$   & $0.84$  & $0.52$  & $64.53$ & $39.68$  \\
        Greg-1  & $0.77$ & $76.25$  & $0.83$ & $0.51$  & $63.61$ & $38.96$ \\
        EagleEye 3G  & $0.78$ & $77.07$  & $0.84$ & $0.53$ & $64.84$  & $40.67$  \\
        \textbf{AdaSAP$_P$}  & $0.77$ & $\mathbf{78.29}$ & $\mathbf{0.84}$ & $\mathbf{0.55}$  & $\mathbf{66.14}$ & $\mathbf{43.22}$  \\
        \bottomrule
        \end{tabular}
    
    \label{tab:resnet-result}
\end{table}

\begin{table}[t]
    \centering
    \footnotesize
    \caption{\textbf{ResNet50 - Latency Reduction.}}
        \begin{tabular}{lccccccc}
        \toprule
        Method & Speedup $\uparrow$ & Size $\downarrow$ & Val  & $R_\text{V2}$ & $R_\text{C}$ & IN-V2 & IN-C \\
        \midrule
        \midrule
        \multicolumn{8}{c}{\textbf{ResNet50}} \\
        Dense & $1$ & $1$ & $77.32$  & $0.838$  & $0.55$ & $64.79$ & $42.46$  \\
        \midrule
        HALP & $2.6$  & $0.43$ & $74.46$  & $0.82$ & $0.47$ & $61.21$ & $35.03$  \\
        \textbf{AdaSAP$_L$ } & $2.6$  & $0.41$  & $\mathbf{75.37}$   & $\mathbf{0.83}$ & $\mathbf{0.50}$ & $\mathbf{62.61}$ & $\mathbf{37.93}$\\
        \midrule
        HALP & $1.6$ & $0.68$ & $76.55$  & $0.83$ & $0.51$ & $63.62$  & $39.13$ \\
        SMCP & $1.7$ & $0.60$ & $76.62$ & $0.83$  & $0.51$  & $63.86$ & $38.72$ \\
        \textbf{AdaSAP$_L$} & $1.6$  & $0.65$  & $\mathbf{77.28}$    & $\mathbf{0.85}$ & $\mathbf{0.54}$ & $\mathbf{65.35}$ & $\mathbf{41.63}$\\
        \midrule
        HALP  & $1.2$ & $0.87$ & $77.45$  & $0.84$  & $0.53$ & $64.88$ & $40.77$ \\
        SMCP& $1.2$ & $0.87$  & $77.57$  &  $0.84$ &$0.53$ & $65.02$ & $40.91$ \\
        \textbf{AdaSAP$_L$}  & $1.1$ &  $0.82$ & $\mathbf{77.93}$   & $0.84$  & $\mathbf{0.55}$ & $\mathbf{65.53}$ & $\mathbf{42.92}$ \\
        \bottomrule
        \end{tabular}
    
    \label{tab:resnet-cmp-halp}
    \vspace{-1.6em}
\end{table}

\textbf{Latency Reduction.}
In a second set of experiments, we focus on latency constraints. 
To this end, we consider the most recent hardware-aware pruning method HALP~(NeurIPS'22~\citep{shen2022structural}), which incorporates latency into the metric to measure the saliency of each neuron.
 Table \ref{tab:resnet-cmp-halp} shows that using AdaSAP$_L$ outperforms prior works both in terms of the $R_\text{C}$ and $R_\text{V2}$ robustness ratios, and also in terms of standard validation performance and performance on OOD datasets.

\subsection{Object Detection}
We show object detection results in Table \ref{tab:obj-det}. 
Across two pruning levels, AdaSAP consistently outperforms the HALP baseline by noticeable margins.
More details can be found in Appendix \ref{appendix_exp}.

\begin{figure}
\begin{minipage}{0.48\linewidth}
\begin{table}[H]
    \vspace{-0.9em}
    \scriptsize
    \centering
    \caption{\textbf{Object Detection.} mAP on validation images and ImageNetC style corruptions on the Pascal VOC dataset.}
        \begin{tabular}{lcccc}
        \toprule
        Method  & Size $\downarrow$ & Val & Corrupted & $R_\text{C}$ \\
        \midrule
        \midrule
        HALP & $0.40$& $0.774$  & $0.583$ & $0.753$ \\
        \textbf{AdaSAP } & $0.40$& $\mathbf{0.795}$ & $\mathbf{0.620}$ & $\mathbf{0.780}$ \\
        \midrule
        HALP & $0.20$& $0.770$ & $0.580$ & $0.753$ \\
        \textbf{AdaSAP } & $0.20$& $\mathbf{0.793}$  & $\mathbf{0.616}$ & $\mathbf{0.776}$\\
            \bottomrule
            \end{tabular}
    
    \label{tab:obj-det}
\end{table}

\vspace{-0.5em}

\begin{table}[H]
    \centering
    \scriptsize
    \caption{\textbf{Sharpness before and after pruning without finetuning}. Lower sharpness values indicates a flatter loss landscape. AdaSAP achieves flatter minima both directly before and after pruning, before any finetuning.}

        \begin{tabular}{lccc}
        \toprule
         & & Sharpness  & Sharpness \\
         Method &  Size $\downarrow$ & pre pruning & post pruning\\
        \midrule
        \midrule
        Taylor &  $0.42$ & $0.039$  & $0.044$   \\
        \textbf{AdaSAP$_P$} &  $\mathbf{0.40}$ & $\textbf{0.037} $  & $\textbf{0.039} $  \\
        \midrule
        Taylor &   $0.76$ & $0.039$ & $0.041$ \\
        \textbf{AdaSAP$_P$} &  $\mathbf{0.76}$ & $\textbf{0.037}$  &  $\textbf{0.038}$ \\
        \bottomrule
        \end{tabular}

    \label{tab:sharpness}
\end{table}

\end{minipage}
\hspace{0.8em}
\begin{minipage}{0.48\linewidth}
\begin{table}[H]
    \vspace{-0.7em}
    \centering
    \scriptsize
    \caption{ \textbf{AdaSAP can work with any pruning criteria.} AdaSAP with Taylor pruning matches or outperforms SGD with Taylor pruning.}

    \label{tab:adasap_taylor}

    \begin{tabular}{lcccc}
    \toprule
        Method  & Size  $\downarrow$ & Val &  $R_\text{C}$   & IN-C \\
        \midrule
        \midrule
        Taylor + SGD & $0.42$ & $75.85$   &  $0.50$  &  $37.84$\\
        AdaSAP$_{P, \text{Taylor}}$ & $0.43$ & $76.26$  &  $0.50$ &  $38.07$\\
        \textbf{AdaSAP$_P$ } & $\textbf{0.41}$ & $\textbf{76.93}$  & $\textbf{0.52}$ &  $\textbf{39.64}$ \\
        \midrule
        Taylor + SGD & $0.76$ & $77.05$  &  $0.52$    &  $39.68$  \\
        AdaSAP$_{P, \text{Taylor}}$ & $0.76$ & $77.42$  & $0.52$  &  $40.27$ \\
        \textbf{AdaSAP$_P$} & $\textbf{0.76}$ & $\textbf{77.86}$   & $\textbf{0.53}$ &  $\textbf{41.30}$  \\
        \bottomrule
    \end{tabular}

\end{table}

\vspace{-2em}
\begin{table}[H]
    \centering
    \scriptsize
    \caption{\textbf{Adaptive perturbations are critical for the best clean and robust performance. }}

    \begin{tabular}{lcccc}
    \toprule
        Method & Size $\downarrow$ & Val & $R_\text{C}$  & IN-C \\
        \midrule
        \midrule
        \multicolumn{5}{c}{\textbf{SGD }} \\
        Dense & $1$&  $77.32$ & $0.54$  & $42.46$\\
        Taylor + SGD  & $0.20$ & $73.56$  &  $0.46$    & $33.93$ \\
        \midrule
        \multicolumn{5}{c}{\textbf{AdaSAP vs. SAM (without ASAM) }} \\
        Taylor + SAM & $0.20$ & $73.62$ &  $0.47$   & $34.49$\\
        \textbf{AdaSAP$_P$} & $0.20$ & $\mathbf{74.38}$ &  $\mathbf{0.48}$     
 &$\mathbf{35.86}$  \\
        \midrule 
        
        \multicolumn{5}{c}{\textbf{AdaSAP vs. SAM (with ASAM) }} \\
        SAM + ASAM & $0.19$& $ 73.93$  &    $0.50$  &  $36.66$ \\
        \textbf{AdaSAP$_P$ + ASAM} &  $0.19$ & $\mathbf{74.63}$  &   $0.50$  &  $\mathbf{37.30}$ \\
        
        \bottomrule
    \end{tabular}
    
    \label{tab:ablation}
\end{table}
\end{minipage}

\end{figure}

\subsection{Ablations}

\noindent
\textbf{Sensitivity to importance metrics.}
In our experiments, we considered $\ell_2$ norm, or magnitude, as the saliency metric for pruning since this led to the best performance. 
However, our method is flexible enough to work with other metrics such as Taylor importance as seen in Table \ref{tab:adasap_taylor}.

\noindent
\textbf{Adaptive perturbations.} 
In order to determine the importance of using adaptive weight perturbations, we compare our method to the use of uniform perturbations throughout all three steps, which is equivalent to using SAM.
In Table~\ref{tab:ablation}, we compare three different setups: Taylor pruning with SGD as a baseline, Taylor pruning with SAM, and AdaSAP$_P$.
 All three settings are fine-tuned for $90$ epochs.
Our method outperforms both SGD and SAM at the same or smaller compressed size. 
This demonstrates that our method's adaptive perturbations during the warmup phase are helping the network achieve a more prunable state. 
See Appendix \ref{appendix_ablations} for more extensive results and further analysis of the choice of uniform perturbations during robustness encouragement.

\subsection{Sharpness Analysis}

Consider the  sharpness defined in ~\citet{foret2020sharpness} as $\max_{\|\eps\|_2 \leq \rho} L_S(\mathbf{w} + \eps) - L_S(\mathbf{w}).$ 
This measures the maximum amount that the loss could change if the weights are perturbed within a ball of radius $\rho$. 
We measure sharpness directly before and after pruning, to evaluate how sharpness prepares the network to be optimally pruned, as well as how pruning affects the model sharpness. Table \ref{tab:sharpness} shows that using AdaSAP leads to flatter models both before and after pruning. In line with previous results that found an association between flatness and generalization, this result can help explain why our models have better generalization and robust generalization performance.

\subsection{Limitations}

One limitation is that AdaSAP requires twice the training time due to the two backward passes necessary for optimization. 
A 90 epoch pruning and finetuning run on 8 V100 GPUs to reduce a ResNet50 network to $0.2 \times$ its size takes takes 32 hours and 12 minutes for AdaSAP$_P$ while SGD with magnitude pruning takes 16 hours and 53 minutes.
Recent strategies ~\citep{du2022sharpness, liu2022towards, jiang2023adaptive} could be used to reduce this additional overhead.
Another limitation is that our analysis is based on existing robustness datasets that don't fully capture the real world corruptions encountered in autonomous driving. 
Despite a focus in the present work on applying pruning (and fine-tuning) to an existing high performing dense model, AdaSAP shows promise in also being used to prune models from scratch.
Finally, although we focus here on the more challenging structured pruning, we believe AdaSAP could also benefit unstructured pruning.

%% file: iclr_2024/conclusion.tex
We introduce the Adaptive Sharpness-Aware Pruning method (AdaSAP) which optimizes for both accuracy and robust generalization during the pruning procedure. 
The method consists of three  steps: application of our novel adaptive weight perturbations, pruning, and flatness-based robustness encouragement. 
AdaSAP outperforms a variety of SOTA pruning techniques on clean and robust performance and relative robustness in both image classification and object detection.

%% file: iclr_2024/appendix.tex
\section{Experiment Details} \label{appendix_exp}
\subsection{Datasets} We conduct image classification experiments on the ImageNet dataset~\citep{deng2009imagenet}. To evaluate the generalization ability of pruned models, we also evaluate models on ImageNet-C~\citep{hendrycks2019benchmarking} and ImageNet-V2~\citep{recht2019imagenet} datasets. The former consists of the validation images from the ImageNet dataset, but with nineteen types of corruptions applied with five different levels of severity. The latter is a dataset created in the same manner as the original dataset, but consisting of different images. Therefore, it is intended to measure out of distribution performance of models. This ImageNet-V2 dataset includes three different sets, each one with a slightly different sampling strategy. We focus our experiments on the~\texttt{MatchedFrequencies} dataset, but include evaluation on all three datasets in Table~\ref{tab:v2_datasets} and see that our method consistently outperforms the baseline. These results do not use the ASAM formulation.

We conduct object detection experiments on the Pascal VOC dataset \citep{everingham2009pascal}. We additionally create a Pascal VOC-C dataset, in which we apply the ImageNet-C corruptions to the Pascal VOC test set. We only use one severity level. 
\subsection{Additional Robustness Datasets}
As described above, we choose the ImageNet V2 \texttt{Matched Frequencies} dataset to perform our main set of experiments, but there are two other ImageNet V2 datasets: \texttt{Threshold 0.7} and \texttt{Top Images}. 
These three datasets vary in terms of selection criteria for included images.
\texttt{Matched Frequencies} attempts to match the image selection frequency of MTurk workers in the original ImageNet validation set.
\texttt{Threshold 0.7} samples from among images with a selection frequency of at least 0.7.
\texttt{Top Images} includes the images with the highest selection frequency within each class.
We provide the pruned model robustness comparison on these the additional ImageNet V2 datasets in Table~\ref{tab:v2_datasets}.


\subsection{Network Architectures}
In our experiments we prune three different networks for the classification task: ResNet50, MobileNet V1 and MobileNet V2 \citep{he2016deep, howard2017mobilenets}. All ablation studies use ResNet50 with ImageNet and ImageNet C datasets unless otherwise specified. ``Val'' column labels refer to standard ImageNet validation accuracy or loss. We use a pretrained model trained for $90$ epochs with a cosine learning rate as in HALP~\citep{shen2022structural} and EagleEye~\citep{li2020eagleeye}. For object detection, we follow the experimental setup in HALP~\citep{shen2022structural} to prune an SSD512 model with ResNet50 as the backbone.
We perform Distributed Data Parallel training across 8 V100 GPUs with batch size 128 for all experiments.

\subsection{Pruning Schedule} Given a pre-trained model, for any architecture, we run the warm up for $10$ epochs, and then we follow the same pruning schedule as in~\cite{de2020progressive}: we prune every $30$ iterations and, in each iteration, we prune away a $p_r$ fraction of neurons so that the final network is pruned by a fraction $p$ (resulting in a network of size $1-p$). To determine the $p_r$ fraction, we follow an exponential decay schedule. Let $k = 1-p$ and let $k_r$ be the number of neurons remaining after $r$ pruning iterations, where the total number of pruning iterations is $R$. Let $m$ be the number of neurons in the dense network, and define $\alpha = \frac{r}{R}$. Then, $k_r = \exp \{ \alpha \log k + (1 - \alpha \log m) \}$.
We fine-tune the pruned model for another $79$ epochs (to reach $90$ epochs total). 

\subsection{Optimization Hyperparameters} The base optimizer is SGD with cosine annealing learning rate with a linear warmup over 8 epochs, a largest learning rate of $1.024$, momentum of $0.875$, and weight decay $3.05e-05$. Unless otherwise stated we use $\rho_{min}=0.01$ and $\rho_{max}=2.0$ for all experiments -- we observe these values scale well across networks and tasks. For robustness encouragement we use $\rho=2.0$ in line with prior work~\citep{kwon2021asam}. 
For some ablation experiments, the original SAM \citep{foret2020sharpness} optimizer is sufficient and in these cases we reduce $\rho_{max}=0.1$ and use constant $\rho=0.05$ for finetuning. 
We include some simple hyperparameter tuning results for $\rho$ values in Table \ref{tab:hyperparam}.

\begin{table}[h]
\centering
\caption{Simple hyperparameter tuning results on ResNet50 model.}
\label{tab:hyperparam}
\begin{tabular}{llcc}
\toprule
$\rho_{min}$ & $\rho_{max}$ & Epoch 2 Val Loss & Epoch 2 Val Acc \\
\midrule
\midrule
$\mathbf{0.01}$ & $\mathbf{0.05}$ & $\mathbf{5.54}$ & $\mathbf{8.63}$ \\
$0.01$ & $0.1$ & $5.59$ & $7.11$ \\
$0.05$ & $0.1$ & $5.73$ & $5.69$ \\
$0.05$ & $0.5$ & $6.92$ & $0.1$ \\
\bottomrule
\end{tabular}
\end{table}

\begin{table}[h]
    \centering
    \caption{Our AdaSAP method outperforms the Taylor pruning baseline on the additional ImageNet V2 datasets. Additionally, the AdaSAP model pruned to $76\%$ outperforms the dense model across all three datasets. Results are on ResNet50 and do not use ASAM.}
    \label{tab:v2_datasets}
    \begin{tabular}{lcccc}
    \toprule
        Method & Size $\downarrow$ & Matched Frequences & Threshold 0.7 & Top Images \\
        \midrule
        \midrule
        Dense & $1$ & $64.80$ & $73.79$ & $79.00$ \\
        \midrule
        Taylor &$0.20$ & $60.56$ & $69.74$ & $75.53$ \\
        \textbf{AdaSAP$_P$} & $0.20$ & $\textbf{62.03}$ & $\textbf{70.83}$ & $\textbf{76.62}$  \\
        \midrule
        Taylor  & $0.40$ & $63.51$ & $72.54$ & $77.83$\\
        \textbf{AdaSAP$_P$} & $0.40$& $\textbf{64.62}$ & $\textbf{73.75}$  & $\textbf{78.87}$ \\
        \midrule
        Taylor & $0.76$& $64.53$ & $73.32$ & $78.72$ \\
        \textbf{AdaSAP$_P$} & $0.76$ & $\textbf{66.00}$ & $\textbf{74.70}$ & $\textbf{79.66}$\\
        \bottomrule
    \end{tabular}
    
 \end{table}

\section{Additional Results} \label{appendix_results}
\subsection{MobileNet Results}

We include results on MobileNet V1 and MobileNet V2 in Tables \ref{tab:mbV1Taylor} and \ref{tab:mobilenet-cmp-halp}. These parallel our main results in the main paper, in which we evaluated AdaSAP$_P$ and AdaSAP$_L$ on ResNet50 and compared them against other pruning methods. Here, we perform a similar comparison, where Table \ref{tab:mbV1Taylor} includes our results on AdaSAP$_P$ and Table \ref{tab:mobilenet-cmp-halp} includes our results on AdaSAP$_L$. We compare against Taylor importance~\citep{molchanov2019importance}, EagleEye ~\citep{li2020eagleeye}, and PolarReg ~\citep{zhuang2020neuron} for AdaSAP$_P$ and against HALP~\citep{shen2022structural}, SMCP~\citep{humble2022soft}, MetaPruning~\citep{liu2019metapruning}, AutoSlim~\citep{yu2019autoslim}, AMC~\citep{he2018amc}, and EagleEye ~\citep{li2020eagleeye} for AdaSAP$_L$. Results here do not use ASAM. We can observe that AdaSAP performs strongly compared to baselines, particularly in the parameter-based setting.

\begin{table}[!b]
    \centering
    \caption{\textbf{MobileNet-V1/V2 - Parameters.} Top1 Accuracy as a function of the pruning ratio.}
    \label{tab:mbV1Taylor}
        \begin{tabular}{lcccccc}
        \toprule
         Method  & Size $\downarrow$ & Val & $R_\text{V2}$ & $R_\text{C}$  & IN-V2  & IN-C  \\
        \midrule
        \midrule
        \multicolumn{7}{c}{\textbf{MobileNet-V1}} \\
        Dense & $1$ & $72.63$ & $0.82$ & $0.45$  & $59.30$ & $32.79$    \\
        \midrule
        Taylor  & $0.40$  & $69.61$ & $0.80$ & $0.42$  & $55.90$  & $29.51$  \\
        \textbf{AdaSAP$_P$ }  & $\mathbf{0.39}$ & $\mathbf{71.05}$ & $\mathbf{0.82}$ & $\mathbf{0.44}$  & $\mathbf{58.08}$ & $\mathbf{31.06}$  \\
        \midrule
        Taylor  & $0.52$ & $71.21$  & $0.81$ & $0.43$ & $57.65$  & $30.92$ \\
        \textbf{AdaSAP$_P$ }  &$\textbf{0.52}$ & $\mathbf{71.58}$  & $\mathbf{0.82}$ & $\mathbf{0.44}$  & $\mathbf{58.75}$ & $\mathbf{31.77}$  \\
        \midrule
        EagleEye & $0.56$ & $70.86$  & $0.80$ & $0.42$   & $56.88$ & $29.98$ \\
        \textbf{AdaSAP$_P$ }  & $\textbf{0.56}$ & $\mathbf{72.05}$ & $\mathbf{0.82}$ & $\mathbf{0.45}$   & $\mathbf{58.94}$ & $\mathbf{32.24}$  \\
        \midrule
        \midrule
        \multicolumn{7}{c}{\textbf{MobileNet-V2}} \\
        Dense  & $1$  & $72.10$ & $0.81$ & $0.45$ & $58.50$ & $32.40$   \\
        \midrule
        Taylor  & $0.72$ & $70.49$  & $0.81$  & $0.43$  & $56.87$ & $30.22$  \\
        \textbf{AdaSAP$_P$ }  & $\textbf{0.72}$ & $\mathbf{72.06}$ & $\mathbf{0.82}$ & $\mathbf{0.45}$  & $\mathbf{58.73}$ & $\mathbf{32.63}$  \\
        \midrule
        PolarReg & $0.87$ & $71.72$ & $0.82$ & $0.45$  & $58.46$ & $32.04$   \\
        Taylor  & $0.88$  & $71.93$ & $0.82$ & $0.45$ & $58.75$  & $32.21$   \\
        \textbf{AdaSAP$_P$ }  & $0.88$ & $\mathbf{72.34}$  & $\mathbf{0.82}$ & $\mathbf{0.45}$  & $\mathbf{59.28}$  & $\mathbf{32.80}$ \\
        
        \bottomrule
        \end{tabular}
\end{table}

\begin{table}[!b]
    \centering
    \caption{\textbf{MobileNet-V1/V2 - Latency.} Top1 accuracy for latency constrained pruning for various speedup ratios. ``--'' indicates that we could not evaluate the model due to unavailable code or models.
    }
        \begin{tabular}{lcccccc}
        \toprule
        Method  & Speedup $\uparrow$ &  Val & $R_\text{V2}$ & $R_\text{C}$  & IN-V2 & IN-C   \\
        \midrule
        \midrule
        \multicolumn{7}{c}{\textbf{MobileNet-V1}} \\
        Dense & $1$ &  $72.63$ & $0.82$ & $0.45$  & $59.30$ & $32.79$   \\
        \midrule
        MetaPruning & $2.06$ &   $66.1$ & -- & -- & -- & --    \\
        AutoSlim & $2.27$ &  $67.9$ & -- & -- & -- & --  \\
        HALP & $2.32$ &  $68.30$  & $0.80$ & $0.41$  & $54.95$ & $28.15$  \\
        SMCP & $\mathbf{2.39}$  &  $68.34$ & $0.80$ & $\textbf{0.42}$  & $54.38$  & $\textbf{28.68}$\\
        \textbf{AdaSAP$_L$}  &   $2.33$ &  $\mathbf{68.45}$ & $\mathbf{0.81}$ & $0.41$  & $\mathbf{55.42}$ & $28.29$   \\
        
        \midrule
        $0.75 $ MobileNetV1 & $1.37$  & $68.4$ & -- & -- & -- & --  \\
        AMC & $1.42$ &   $70.5$ & -- & -- & -- & --  \\
        MetaPruning  & $1.42$  & $70.9$ & -- & -- & -- & --  \\
        EagleEye & $1.47$  & $70.86$ & $0.80$ & $0.42$  & $56.88$ & $29.98$ \\
        HALP & $1.68$ &  $71.31$ & $0.81$ & $0.43$  & $57.38$ & $30.77$  \\
        SMCP  & $\mathbf{1.72}$  & $71.00$ & $0.81$ & $0.44$  & $57.20$ & $31.02$  \\
        \textbf{AdaSAP$_L$}  & $1.70$  & $\mathbf{71.48}$ & $\mathbf{0.82}$ & $\mathbf{0.44}$ & $\mathbf{58.23}$  & $\mathbf{31.35}$  \\
        \midrule

        \midrule
        \multicolumn{7}{c}{\textbf{MobileNet-V2}} \\
        Dense & $1$ &  $72.10$ & $0.81$ & $0.45$  & $58.50$ & $32.40$   \\
        \midrule
        HALP  & $\mathbf{1.84}$  & ${70.42}$ & $0.81$ & $0.45$  & $57.21$  & $31.69$ \\
        \textbf{AdaSAP$_L$}  & $1.81$  & $\mathbf{71.35}$ & $0.81$ & $\mathbf{0.46}$  & $\mathbf{57.85}$ & $\mathbf{32.63}$  \\
        \midrule
        HALP  & $1.33$ &  $72.16$  & $0.81$ & $0.46$ & $58.53$ & $\mathbf{33.04}$   \\
        \textbf{AdaSAP$_L$}  & $\mathbf{1.39}$  &  $\mathbf{72.19}$ & $\mathbf{0.82}$ & $0.46$ & $\mathbf{59.36}$& $32.91$  \\
        \bottomrule
        \end{tabular}
    
    \label{tab:mobilenet-cmp-halp}
\end{table}

\subsection{Margin of Improvement in Classification}

\begin{figure}[t]
    \centering
    \includegraphics[width=0.8\linewidth]{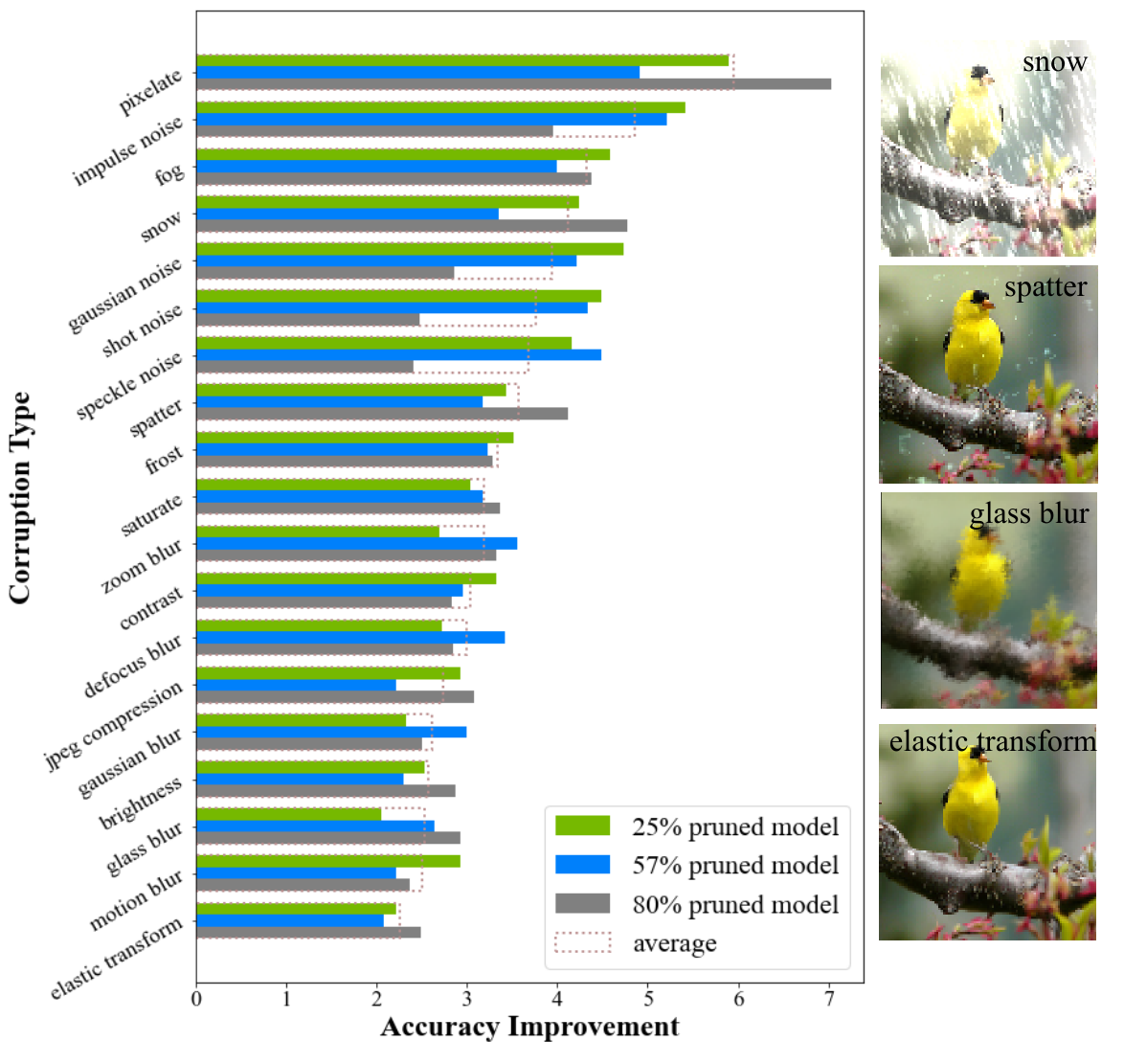}
    \caption{\textbf{ImageNet C Dataset.} Performance difference on various ImageNet C corruption types on models of varying sparsity. Accuracy improvement is the Top1 accuracy on a ResNet50 model trained and pruned using AdaSAP minus that of a Taylor pruned model.}
    \label{fig:margin}
\end{figure}

Figure \ref{fig:margin} shows the margin of performance on various corruption types for the classification task. AdaSAP outperforms a Taylor pruned model on all corruptions, across three different sparsities. We can see that it tends to particularly improve performance on several corruptions that may be important for the autonomous driving application, such as pixelated images, fog, and snow.

\subsection{Margin of Improvement in Object Detection}
\begin{figure}[t]
    \centering
    \includegraphics[width=0.65\linewidth]{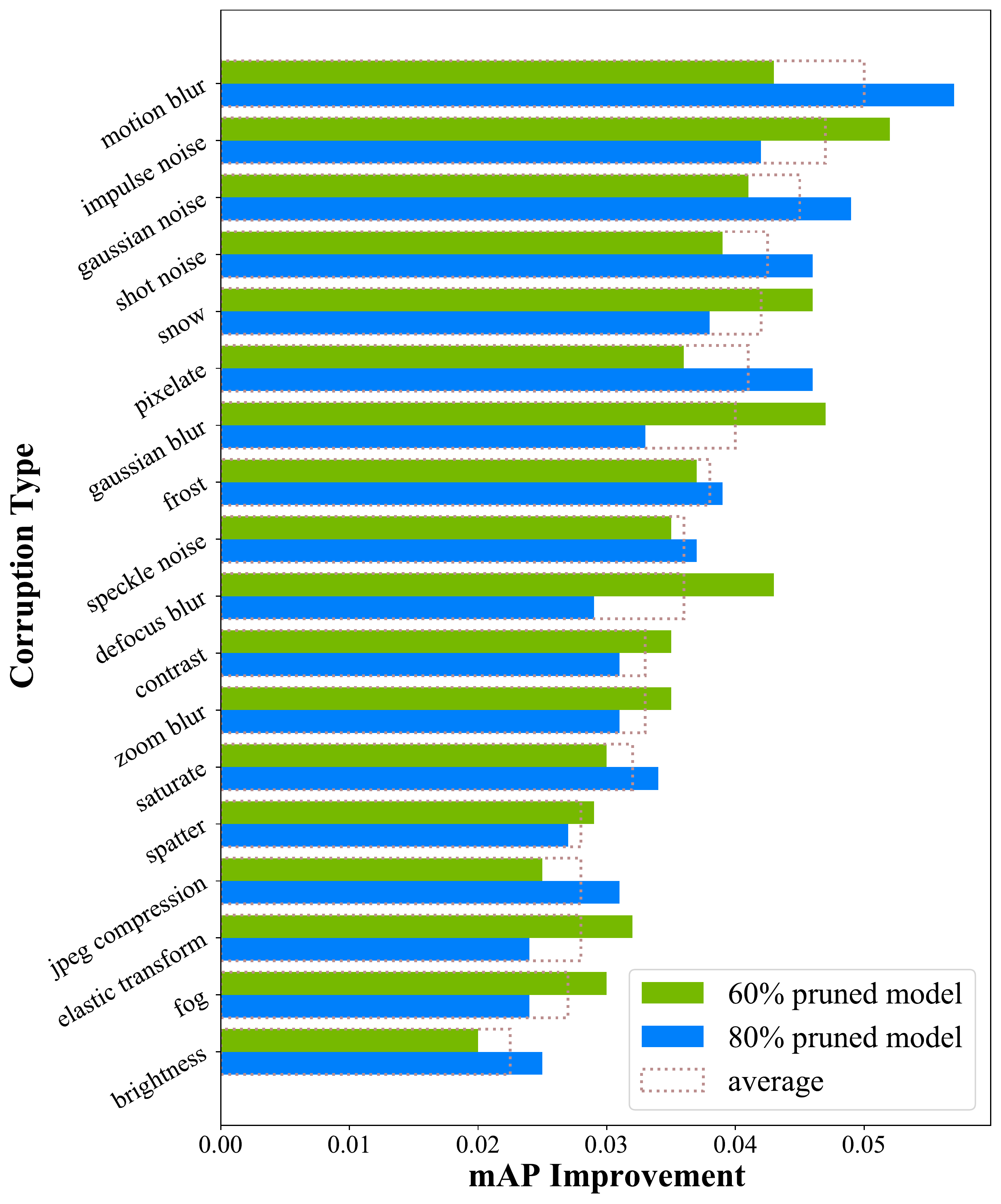}
    \caption{\textbf{Pascal VOC-C dataset.} Performance difference on various ImageNet C corruption types on models of varying sparsity. mAP improvement is the mAP of a ResNet50 model trained and pruned using AdaSAP minus that of a HALP pruned model.}
    \label{fig:margin_od}
\end{figure}

Similarly to our result on classification, we include margins of performance improvement on various corruption types for the object detection task in Figure \ref{fig:margin_od}.

\subsection{Relative Robustness}
\begin{figure}[t]
    \centering
    \includegraphics[width=\linewidth]{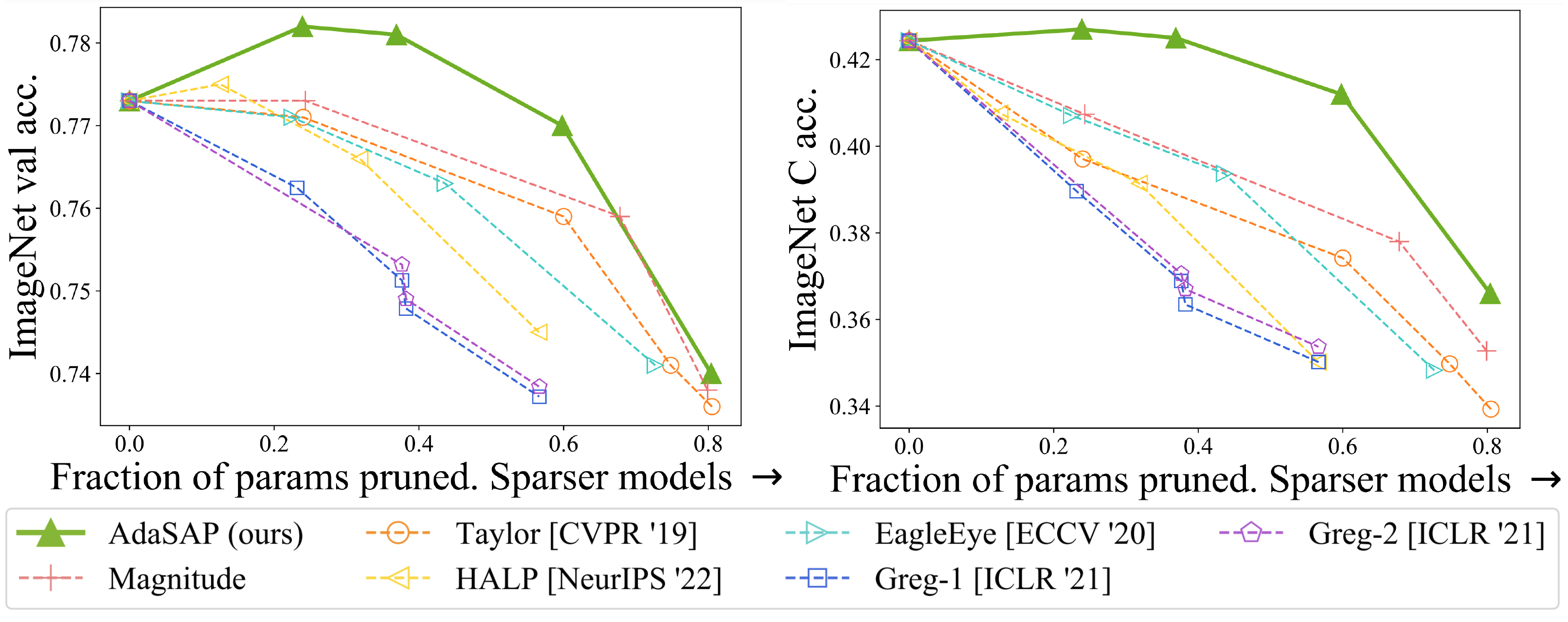}
    \caption{\textbf{ImageNet Validation and ImageNet C performance on AdaSAP vs. baselines.} Contains the same data as used to produce Figure \ref{fig:fig1} but demonstrates that AdaSAP additionally dominates baselines on both components of the relative robustness metric. }
    \label{fig:separate_fig1}
\end{figure}

In Figure \ref{fig:separate_fig1} we show that AdaSAP outperforms baselines on each of the constituent elements of the relative robustness metric. 
Recall that relative robustness is robust accuracy divided by standard validation accuracy.
In addition to outperforming baselines on relative robustness, AdaSAP also outperforms on ImageNet validation accuracy and ImageNet C robust accuracy.

\subsection{Additional Sharpness Metrics}
We additionally run limited measurements of an alternative sharpness metric: the top eigenvalue of the Hessian. We compare the sharpness of a model trained with AdaSAP vs. SGD in Table \ref{tab:hessian_sharpness}.

\begin{table}[h]
\centering
\caption{Top Hessian eigenvalue sharpness measurements on ResNet50 model.}
\label{tab:hessian_sharpness}
\begin{tabular}{llcc}
\toprule
Optimizer & Pre-pruning & Post-pruning & Post-finetuning \\
\midrule
\midrule
SGD & $35.91$& $93.82$ & $8.17$ \\
\textbf{AdaSAP$_L$} & $\mathbf{25.23}$& $\mathbf{16.79}$ &$\mathbf{7.66}$ \\
\bottomrule
\end{tabular}
\end{table}

\begin{table}[!t]
    \centering
    \caption{ \textbf{Sensitivity to pruning criteria.} AdaSAP performs best when combined with magnitude pruning, but is flexible enough to be used with other criteria, such as Taylor importance. Here we see that AdaSAP with Taylor pruning matches or outperforms SGD with Taylor pruning. Results are on ResNet50 and do not use ASAM.}
    \begin{tabular}{lcccccc}
    \toprule
        Method  & Size  $\downarrow$ & Val & $R_\text{V2}$ & $R_\text{C}$   & IN-V2& IN-C \\
        \midrule
        \midrule
        Taylor + SGD   & $0.42$ & $75.85$   & $0.84$ & $0.50$  & $63.51$ & $37.84$\\
        AdaSAP$_{P, \text{Taylor}}$  & $0.43$ & $76.26$  & $0.84$ & $0.50$ & $63.77$ & $38.07$\\
        \textbf{AdaSAP$_P$ } & $\textbf{0.41}$ & $\textbf{76.93}$ & $\textbf{0.84}$ & $\textbf{0.52}$ & $\textbf{64.49}$  & $\textbf{39.64}$ \\
        \midrule
        Taylor + SGD & $0.76$ & $77.05$  & $0.84$  & $0.52$    & $64.53$ & $39.68$  \\
        AdaSAP$_{P, \text{Taylor}}$   & $0.76$ & $77.42$  & $0.84$ & $0.52$  & $65.24$ & $40.27$ \\
        \textbf{AdaSAP$_P$ } & $\textbf{0.76}$ & $\textbf{77.86}$ & $\textbf{0.85}$  & $\textbf{0.53}$ & $\textbf{66.00}$ & $\textbf{41.30}$  \\
        \bottomrule
    \end{tabular}

    \label{tab:adasap_taylor_complete}
\end{table}

\begin{table}[!t]
\centering
\caption{\textbf{Confidence intervals over three repeats.} EagleEye checkpoints are obtained from the official repository with only one seed. Results are on ResNet50.}

\begin{tabular}{lcccc}
    \toprule
    Method  & Size $\downarrow$ & Val & IN-V2 & IN-C  \\
    \midrule
    \midrule
    Magnitude  &  $0.20$ & $73.71 \pm 0.12$  & $61.21 \pm 0.17$ & $35.33 \pm 0.18$  \\
    Taylor & $0.20$ & $73.42 \pm 0.19$   & $60.36 \pm 0.19$ & $33.97 \pm 0.12$ \\
    EagleEye  & $0.27$ & $74.13$   & $61.30$ & $34.84$ \\
    \textbf{AdaSAP$_P$ }  & $0.20$ & $\mathbf{74.54 \pm 0.09}$  & $\mathbf{62.21 \pm 0.13}$ & $\mathbf{37.30 \pm 0.19}$   \\
    \midrule 
    Magnitude  & $0.76$ & $77.32 \pm 0.06$   & $65.18 \pm 0.27$  &$40.64 \pm 0.20$ \\
    Taylor  & $0.76$ & $77.05$  & $64.53$ & $39.68$ \\
    EagleEye  & $0.78$ & $77.07$  & $64.84$ & $40.67$ \\
    \textbf{AdaSAP$_P$ }  & $0.77$ & $\mathbf{78.23 \pm 0.06}$ & $\mathbf{65.98 \pm 0.32}$ & $\mathbf{43.43 \pm 0.20}$ \\
    \bottomrule
\end{tabular}
\label{confidence_intervals_complete}
\end{table}

\section{Additional Ablations} \label{appendix_ablations}
\subsection{Performance change after pruning}
In Table \ref{tab:val_prune_acc_before_after} we show how the loss and accuracy change over the course of pruning. 
Our hypothesis is that our method sets up the network for better pruning, so that the performance drop over the course of pruning is minimized.
In most cases, our method has the best validation loss and accuracy both before and after pruning.
This indicates that our method sets up the model to be pruned well, and also preserves performance well throughout the pruning process.

\begin{table}[t]
    \centering
    \caption{Validation Loss and Accuracy directly before and after pruning (before additional fine-tuning).  Across several pruning levels, our method generally reaches the lowest validation loss and accuracy both before and after pruning. Results are on ResNet50 and do not use ASAM.}
    \begin{tabular}{lccccc}
    \toprule
        Method  & Size $\downarrow$ & Val Loss Before & Val Loss After & Val Acc Before & Val Acc After \\
        \midrule
        \midrule
        Taylor  & $0.20$ & $\textbf{2.245}$   &  $3.315$ & $\textbf{67.798}$ & $42.915$ \\
        Mag  & $0.20$ & $2.416$ &   $3.288$ & $63.843$ & $43.21$\\
        SAM  & $0.20$ & $2.255$ &    $3.2$ & $67.577$ & $45.802$\\
        AdaSAP$_P$   & $0.20$ &  $2.293$  &    $\textbf{3.146}$ & $66.555$  & $\textbf{46.313}$  \\
        \midrule
        Taylor  & $0.43$ & $2.284$    & $2.69$ & $66.67$ & $57.046$ \\
        Mag   & $0.43$& $2.408$ &   $2.587$ & $63.844$ & $59.275$\\
        SAM   & $0.43$& $2.327$ &    $2.676$ & $65.541$ & $57.449$\\
        AdaSAP$_P$    &$0.43$ & $\textbf{2.251}$  & $\textbf{2.574}$ & $\textbf{67.473}$  & $\textbf{59.962}$  \\
        \midrule
        Taylor   & $0.76$ & $2.441$   &  $2.42$ & $63.389$ & $63.711$ \\
        Mag  & $0.76$ & $2.292$ &   $2.406$ & $66.714$ & $64.087$\\
        SAM   & $0.76$& $2.401$ &    $2.379$ & $64.019$ & $64.589$\\
        AdaSAP$_P$    & $0.76$& $\textbf{2.213}$  &   $\textbf{2.347}$ & $\textbf{68.463}$  & $\textbf{65.393}$  \\
        \bottomrule
    \end{tabular}
    
    \label{tab:val_prune_acc_before_after}
\end{table}

\subsection{Varying the length of the adaptive weight perturbation step.}
Throughout the main set of experiments use 10 epochs for the adaptive weight perturbation step. This delineates how long we use the AdaSAP optimizer for, as well as how soon into the procedure we begin pruning. In this experiment, we analyze the sensitivity of the approach to this parameter. We report results for this experiment in Table~\ref{tab:vary_warmup}. We can observe that as we make this period longer, validation accuracy and ImageNet C accuracy both drop slightly, while ImageNet V2 seems to have no discernible pattern. Ratios $R_C$ and $R_{V2}$ also stay consistent across the experiment. 

\begin{table}[!t]
    \centering
    \caption{ \textbf{Effects of varying number of epochs of adaptive weight perturbation.} Increasing the number of epochs leads to smaller models but slightly worse validation and ImageNet C performance. Results are on ResNet50.}
    \begin{tabular}{lcccccc}
    \toprule
        Num epochs & Size $\downarrow$ & Val & $R_\text{V2}$ & $R_\text{C}$ & IN-V2 & IN-C  \\
        \midrule
        \midrule
        $5$  & $0.48$ & $73.89$ & $0.83$ & $0.48$  & $61.29$ & $35.65$  \\
        $10$ & $0.46$ & $73.82$ & $0.83$ & $0.48$ & $60.91$ & $35.59$  \\
        $20$ & $0.44$ & $73.63$ & $0.84$ & $0.48$ & $61.48$ & $35.25$  \\
        $30$ & $0.43$ & $73.47$ & $0.83$ & $0.48$ & $60.72$ & $35.04$  \\
        \bottomrule
    \end{tabular}
    
    \label{tab:vary_warmup}
\end{table}
\begin{table}[t!]
    \centering
    \caption{Comparison of various weight perturbation strategies during robustness encouragement. Results are on ResNet50.}
    \begin{tabular}{lccc}
    \toprule
    Perturbation Type & Val & IN-C & IN-V2 \\
    \midrule
    \midrule
    No weight perturbations & $73.98$ & $35.84$ & $62.00$ \\
    Adaptive weight perturbations & $74.10$ & $35.72$ & $61.94$ \\
    Uniform weight perturbations & $\mathbf{74.39}$ & $\mathbf{35.86}$ & $\mathbf{62.03}$ \\
    \bottomrule
\end{tabular}
    
    \label{diff_robustness_encouragement_opt}
\end{table}

\subsection{Robustness Encouragement}
As mentioned in the main text, we consider various ablations to determine the necessity of various steps of the AdaSAP procedure. In the third step of our procedure, robustness encouragement, we choose to apply uniform perturbations across all weights in the network. This differs from the first step, in which we apply adaptive weight perturbations. In Table \ref{diff_robustness_encouragement_opt}, we examine the effects of different weight perturbation strategies during the robustness encouragement phase. We can see that while all three strategies lead to relatively close final performance across the three datasets, uniform weight perturbations perform slightly better, suggesting that our choice of applying them in our procedure may be slightly benefitting the performance.

\subsection{Importance of Adaptive Weight Perturbations}

In Table \ref{tab:ablation_appendix} we extend an ablation from the main paper in which we compare AdaSAP to SAM, effectively evaluating the importance of warmup with adaptive weight perturbations. Here, we perform the comparison on a wider range of sparsities and observe that a similar pattern emerges.

\begin{table}[!t]
    \centering
        \caption{\textbf{Comparison of AdaSAP to SAM optimizer.} AdaSAP outperforms SAM and SGD on standard validation performance and ImageNet-C performance, but slightly trails SAM on ImageNet-V2 when both methods are augmented with ASAM. Results are on ResNet50.
    }
    \begin{tabular}{lcccccc}
    \toprule
        Method  & Size $\downarrow$ & Val & $R_\text{V2}$  & $R_\text{C}$  &  IN-V2 & IN-C \\
        \midrule
        \midrule
        \multicolumn{7}{c}{\textbf{SGD }} \\
        Dense & $1$&  $77.32$ & $0.84$ & $0.54$ &   $64.79$  & $42.46$\\
        \midrule
        Taylor + SGD  & $0.20$ & $73.56$  & $0.82$ & $0.46$   & $60.56$ & $33.93$ \\
        \midrule
        \multicolumn{7}{c}{\textbf{AdaSAP vs. SAM (without ASAM) }} \\
        Taylor + SAM  & $0.20$ & $73.62$ & $0.83$  & $0.47$   & $61.37$ & $34.49$\\
        \textbf{AdaSAP$_P$ }  & $0.20$ & $\mathbf{74.38}$ & $\mathbf{0.83}$ & $\mathbf{0.48}$    & $\mathbf{62.03}$ 
 &$\mathbf{35.86}$  \\
        \midrule 
        Taylor + SGD  & $0.42$ & $75.85$  & $0.84$ & $0.50$  & $63.51$  & $37.84$ \\
        Taylor + SAM   & $ 0.43$  & $76.27$ & $0.84$ & $0.50$    & $63.81$ & $37.72$\\
        \textbf{AdaSAP$_P$ }  & $\mathbf{0.40}$  & $\mathbf{77.03}$ & $\mathbf{0.84}$ & $\mathbf{0.51}$  & $\mathbf{64.62}$  & $\mathbf{39.57}$\\
        \midrule 
        Taylor + SGD  & $0.76$& $77.05$ & $0.84$    & $0.52$    & $64.53$ & $39.68$   \\
        Taylor + SAM   & $0.76$ & $77.29$  & $0.84$ & $0.52$ & $64.84$ & $40.07$ \\
        \textbf{AdaSAP$_P$ } & $0.76$ & $\mathbf{77.86}$ & $\mathbf{0.85}$ & $\mathbf{0.53}$   & $\mathbf{66.00}$ & $\mathbf{41.30}$  \\
        \midrule
        \multicolumn{7}{c}{\textbf{AdaSAP vs. SAM (with ASAM) }} \\
        SAM + ASAM  & $0.19$& $ 73.93$  &  $0.84$  &  $0.50$  & $61.76$ & $36.66$ \\
        \textbf{AdaSAP$_P$ + ASAM } &  $0.19$ & $\mathbf{74.63}$  & $0.83$ &  $0.50$  & $\mathbf{62.08}$  & $\mathbf{37.30}$ \\
        \midrule 
        SAM + ASAM  &  $0.41$ & $77.10$ & $0.84$  & $0.53$ & $64.94$   & $ 40.90$\\
        \textbf{AdaSAP$_P$ + ASAM } &  $\mathbf{0.40}$ & $\mathbf{77.27}$ & $0.83$ & $0.53$   & $64.51$ & $\mathbf{41.23}$\\
        \bottomrule
    \end{tabular}

    \label{tab:ablation_appendix}
\end{table}